\documentclass[11pt]{article}
\usepackage{coling2016}
\usepackage{times}
\usepackage{url}
\usepackage{latexsym}
\usepackage{amsfonts}
\usepackage{amssymb}
\usepackage{tikz}
\usepackage{standalone}
\usepackage{subcaption}
\usepackage{pgfplots}

\title{Neural-based Noise Filtering from Word Embeddings}

\author{Kim Anh Nguyen \and Sabine Schulte im Walde \and Ngoc Thang Vu \\
	    Institut f\"ur Maschinelle Sprachverarbeitung\\
	    Universit\"at Stuttgart\\
	    Pfaffenwaldring 5B, 70569 Stuttgart, Germany\\
	    {\{\tt nguyenkh,schulte,thangvu\}@ims.uni-stuttgart.de}}

\date{}

\begin{document}
\maketitle

\begin{abstract}
Word embeddings have been demonstrated to benefit NLP tasks impressively. Yet, there is room for improvement in the vector representations, because current word embeddings typically contain unnecessary information, i.e., \textit{noise}. We propose two novel models to improve word embeddings by unsupervised learning, in order to yield word denoising embeddings. The word denoising embeddings are obtained by strengthening salient information and weakening noise in the original word embeddings, based on a deep feed-forward neural network filter. Results from benchmark tasks show that the filtered word denoising embeddings outperform the original word embeddings. 
\end{abstract}

\section{Introduction}
\label{sec:intro}
Word embeddings aim to represent words as low-dimensional dense vectors. In comparison to distributional count vectors, word embeddings address the problematic sparsity of word vectors and achieved impressive results in many NLP tasks such as sentiment analysis (e.g., \newcite{Kim:14}), word similarity (e.g., \newcite{Pennington:14}), and parsing (e.g., \newcite{Lazaridou:13}). Moreover, word embeddings are attractive because they can be learned in an unsupervised fashion from unlabeled raw corpora. There are two main approaches to create word embeddings. The first approach makes use of neural-based techniques to learn word embeddings, such as the Skip-gram model~\cite{Mikolov:13}. The second approach is based on matrix factorization~\cite{Pennington:14}, building word embeddings by factorizing word-context co-occurrence matrices.

In recent years, a number of approaches have focused on improving word embeddings, often by integrating lexical resources. For example, \newcite{Adel/Schuetze:14} applied coreference chains to Skip-gram models in order to create word embeddings for antonym identification. \newcite{Pham:15} proposed an extension of a Skip-gram model by integrating synonyms and antonyms from WordNet. Their extended Skip-gram model outperformed a standard Skip-gram model on both general semantic tasks and distinguishing antonyms from synonyms. In a similar spirit, \newcite{Nguyen:16} integrated distributional lexical contrast into every single context of a target word in a Skip-gram model for training word embeddings. The resulting word embeddings were used in similarity tasks, and to distinguish between antonyms and synonyms. \newcite{Faruqui:15} improved word embeddings without relying on lexical resources, by applying ideas from sparse coding to transform dense word embeddings into sparse word embeddings. The dense vectors in their models can be transformed into sparse overcomplete vectors or sparse binary overcomplete vectors. They showed that the resulting vector representations were more similar to interpretable features in NLP and outperformed the original vector representations on several benchmark tasks.

In this paper, we aim to improve word embeddings by reducing their noise. The hypothesis behind our approaches is that word embeddings contain unnecessary information, i.e. \textit{noise}. We start out with the idea of learning word embeddings as suggested by~\newcite{Mikolov:13}, relying on the distributional hypothesis~\cite{Harris1954} that words with similar distributions have related meanings. We address those distributions in embedded vectors of words that decrease the value of such vector representations. For instance, consider the sentence \textit{the quick brown fox gazing at the cloud jumped over the lazy dog}. The context \textit{jumped} can be used to predict the words \textit{fox}, \textit{cloud} and \textit{dog} in a window size of 5 words; however, a \textit{cloud} cannot \textit{jump}. The context \texttt{jumped} is therefore considered as noise in the embedded vector of \texttt{cloud}. We propose two novel models to smooth word embeddings by filtering noise: We strengthen salient contexts and weaken unnecessary contexts.

The first proposed model is referred to as \textit{complete word denoising embeddings model (CompEmb)}. Given a set of original word embeddings, we use a filter to learn a denoising matrix, and then project the set of original word embeddings into this denoising matrix to produce a set of complete word denoising embeddings. The second proposed model is referred to as \textit{overcomplete word denoising embeddings model (OverCompEmb)}. We make use of a sparse coding method to transform an input set of original word embeddings into a set of overcomplete word embeddings, which is considered as the ``overcomplete process''. We then apply a filter to train a denoising matrix, and thereafter project the set of original word embeddings into the denoising matrix to generate a set of overcomplete word denoising embeddings. The key idea in our models is to use a filter for learning the denoising matrix. The architecture of the filter is a feed-forward, non-linear and parameterized neural network with a fixed depth that can be used to learn the denoising matrices and reduce noise in word embeddings. Using state-of-the-art word embeddings as input vectors, we show that the resulting word denoising embeddings outperform the original word embeddings on several benchmark tasks such as word similarity and word relatedness tasks, synonymy detection and noun phrase classification. Furthermore, the implementation of our models is made publicly available\footnote{\url{https://github.com/nguyenkh/NeuralDenoising}}.

The remainder of this paper is organized as follows: Section~\ref{sec:learning-approx} presents the two proposed models, the loss function, and the sparse coding technique for overcomplete vectors. In Section~\ref{sec:experiments}, we demonstrate the experiments on evaluating the effects of our word denoising embeddings, tuning hyperparameters, and we analyze the effects of filter depth. Finally, Section~\ref{sec:conclusion} concludes the paper.

\section{Learning Word Denoising Embeddings}
\label{sec:learning-approx}
\begin{figure}[t]
	\centering
	\includegraphics[width=\textwidth]{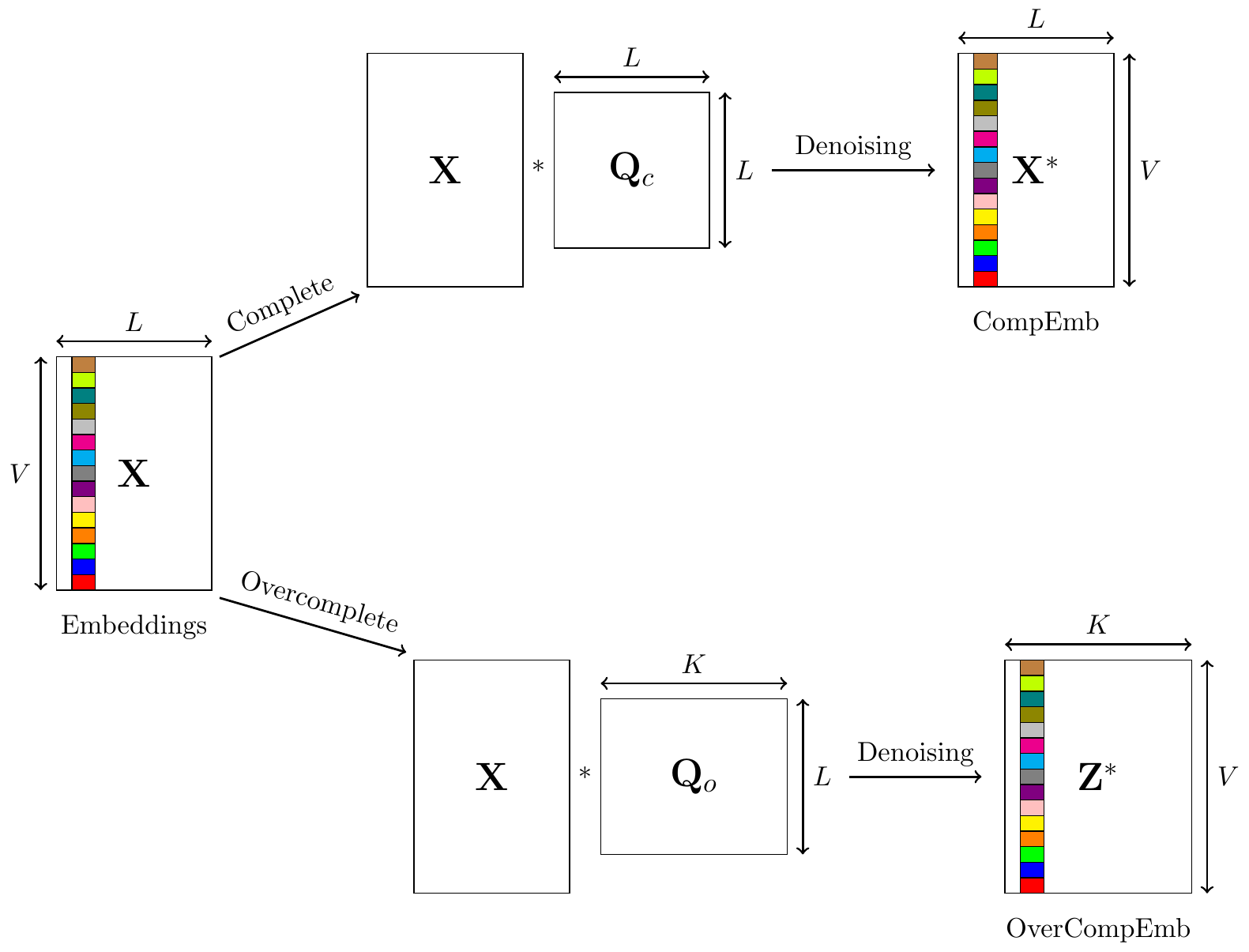} %
	\caption{Illustration of word denoising embeddings methods, with complete word denoising embeddings at the top, and overcomplete word denoising embeddings at the bottom.}
	\label{fig:graph}
        \vspace{+5mm}
\end{figure}

In this section, we present the two contributions of this paper. Figure~\ref{fig:graph} illustrates our two models to learn denoising for word embeddings. The first model on the top, the complete word denoising embeddings model ``CompEmb'' (Section~\ref{subsec:complete-representation}), filters noise from word embeddings $\mathbf{X}$ to produce complete word denoising embeddings $\mathbf{X^*}$, in which the vector length of $\mathbf{X^*}$ in comparison to $\mathbf{X}$ is unchanged after denoising (called \textit{complete}). The second model at the bottom of the figure, the overcomplete word denoising embeddings model ``OverCompEmb'' (Section~\ref{subsec:overcomplete-representation}), filters noise from word embeddings $\mathbf{X}$ to yield overcomplete word denoising embeddings $\mathbf{Z^*}$, in which the vector length of $\mathbf{Z^*}$ tends to be greater than the vector length of $\mathbf{X}$ (called \textit{overcomplete}). 

For the notations, let $\mathbf{X} \in \mathbb{R}^{V \times L}$ is an input set of word embeddings in which $V$ is the vocabulary size, and $L$ is the vector length of $\mathbf{X}$. Furthermore, $\mathbf{Z} \in \mathbb{R}^{V \times K}$ is the overcomplete word embeddings in which $K$ is the vector length of $\mathbf{Z}$ ($K>L$); finally, $\mathbf{D} \in \mathbb{R}^{L \times L}$ is the pre-trained dictionary (Section~\ref{subsec:sparse-coding}).

\subsection{Complete Word Denoising Embeddings}
\label{subsec:complete-representation}
In this subsection, we aim to reduce noise in the given input word embeddings $\mathbf{X}$ by learning a denoising matrix $\mathbf{Q_c}$. The complete word denoising embeddings $\mathbf{X^*}$ are then generated by projecting $\mathbf{X}$ into $\mathbf{Q_c}$. More specifically, given an input $\mathbf{X} \in \mathbb{R}^{V \times L}$, we seek to optimize the following objective function:
\begin{equation}
 \mathop \mathrm{argmin} \limits_{\mathbf{X,Q_c,S}} \sum\limits_{i = 1}^V \left\| \mathbf{x_i} - f(\mathbf{x_i,Q_c,S}) \right\| + \alpha {\left\| \mathbf{S} \right\|_1}
 \label{eq:ObjComplete}
\end{equation}
where $f$ is a filter; $\mathbf{S}$ is a lateral inhibition matrix; and $\alpha$ is a regularization hyperparameter. Inspired by studies on sparse modeling, the matrix $\mathbf{S}$ is chosen to be symmetric and has zero on the diagonal.

\begin{figure*}[t]
	\centering
	\begin{subfigure}{.45\textwidth}
		\centering
		\includegraphics[width=.95\textwidth]{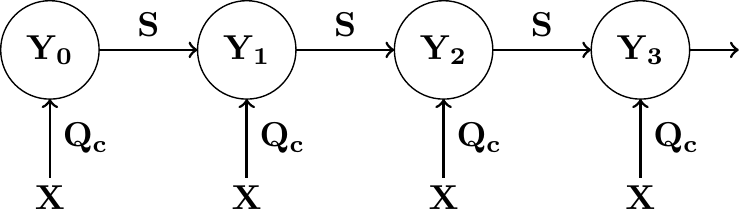} %
		\caption{CompEmb}
		\label{subfig:CompleteRep}
	\end{subfigure}
	\hspace*{+8mm}
	\begin{subfigure}{.45\textwidth}
		\centering
		\includegraphics[width=.95\textwidth]{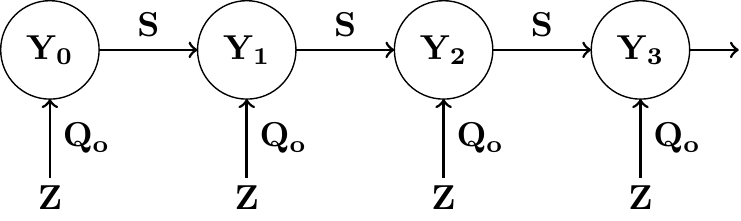} %
		\caption{OverCompEmb}
		\label{subfig:OvercompleteRep}
	\end{subfigure}
	\caption{Architecture of the filters with the fixed depth $T = 3$.}
        \vspace{+5mm}
	\label{fig:architecture}
\end{figure*}

\noindent The goal of this matrix is to implement excitatory interaction between neurons, and to increase the convergence speed of the neural network~\cite{Szlam:11}. More concretely, the matrices $\mathbf{Q_c}$ and $\mathbf{S}$ are initialized with $\mathbf{I}$ and $E$, which are identity matrices, and the Lipschitz constant:
\[
	\begin{array}{l}
		\mathbf{Q_c} = \frac{1}{E} \mathbf{D} \mbox{; }
		\mathbf{S} = \mathbf{I} - \frac{1}{E} \mathbf{D^T} \mathbf{D} \\
		E > \mbox{ the largest eigenvalue of } \mathbf{D^T D} \\
		\mathbf{D} \in \mathbb{R}^{L \times L} \mbox{ be pre-trained dictionary} 
	\end{array}	
\]
The underlying idea for reducing noise is to make use of a filter $f$ to learn a denoising matrix $\mathbf{Q_c}$; hence, we design the filter $f$  as a non-linear, parameterized, feed-forward architecture with a fixed depth that can be trained to approximate $f(\mathbf{X,Q_c,S})$ to $\mathbf{X}$ as in Figure~\ref{subfig:CompleteRep}. As a result, noise from word embeddings will be filtered by layers of the filter $f$. The filter $f$ is encoded as a recursive function by iterating over the number of fixed depth $T$, as the following recursive Equation~\ref{eq:architecture} shows:
\begin{equation}
	\begin{array}{l}
		\mathbf{Y} = f(\mathbf{X,Q_c,S}) \\
		\mathbf{Y}(0) = \mathcal{G}(\mathbf{XQ_c}) \\
		\mathbf{Y}(k+1) = \mathcal{G}(\mathbf{XQ_c} + \mathbf{Y}(k) \mathbf{S}) \\
		0 \leq k < T
	\end{array}
	\label{eq:architecture}
\end{equation} 
$\mathcal{G}$ is a non-linear activation function. The matrices $\mathbf{Q_c}$ and $\mathbf{S}$ are learned to produce the lowest possible error in a given number of iterations. Matrix $\mathbf{S}$, in the architecture of filter $f$, acts as a controllable matrix to filter unnecessary information on embedded vectors, and to impose restrictions on further reducing the computational burden (e.g., solving low-rank approximation problem or keeping the number of terms at zero~\cite{Gregor/LeCun:10}). Moreover, the initialization of the matrices $\mathbf{Q_c}$, $\mathbf{S}$ and $E$ enhances a highly efficient minimization of the objective function in Equation~\ref{eq:ObjComplete}, due to the pre-trained dictionary $\mathbf{D}$ that carries the information of reconstructing $\mathbf{X}$.

The architecture of the filter $f$ is a recursive feed-forward neural network with the fixed depth $T$, so the number of $T$ plays a significant role in controlling the approximation of $\mathbf{X^*}$. The effects of $T$ will be discussed later in Section~\ref{subsec:depth-effects}.
When $\mathbf{Q_c}$ is trained, the complete word denoising embeddings $\mathbf{X^*}$ are yielded by projecting $\mathbf{X}$ into $\mathbf{Q_c}$, as shown by the following Equation~\ref{eq:transform_x}:
\begin{equation}
	\mathbf{X^*} = \mathcal{G}(\mathbf{XQ_c})
	\label{eq:transform_x}
\end{equation}

\subsection{Overcomplete Word Denoising Embeddings}
\label{subsec:overcomplete-representation}
Now we introduce our method to reduce noise and overcomplete vectors in the given input word embeddings. To obtain overcomplete word embeddings, we first use a sparse coding method to transform the given input word embeddings $\mathbf{X}$ into overcomplete word embeddings $\mathbf{Z}$. Secondly, we use overcomplete word embeddings $\mathbf{Z}$ as the intermediate word embeddings to optimize the objective function: A set of input word embeddings $\mathbf{X} \in \mathbb{R}^{V \times L}$ is transformed to overcomplete word embeddings $\mathbf{Z} \in \mathbb{R}^{V \times K}$ by applying sparse coding method in Section~\ref{subsec:sparse-coding}. We then make use of the pre-trained dictionary $\mathbf{D} \in \mathbb{R}^{L \times K}$ and $\mathbf{Z} \in \mathbb{R}^{V \times K}$ to learn the denoising matrix $\mathbf{Q_o}$ by minimizing the following Equation~\ref{eq:ObjOvercomplete}:
\begin{equation}
 	\mathop \mathrm{argmin} \limits_{\mathbf{X,Q_o,S}} \sum\limits_{i = 1}^V \left\| \mathbf{z_i} - f(\mathbf{x_i,Q_o,S}) \right\| + \alpha {\left\| \mathbf{S} \right\|_1}
 \label{eq:ObjOvercomplete}
\end{equation}
The initialization of the parameters $\mathbf{Q_o}$, $\mathbf{S}$, $E$ and $\alpha$ follows the same procedure as described in Section~\ref{subsec:complete-representation}, and with the same interpretation of the filter architecture in Figure~\ref{subfig:OvercompleteRep}. The overcomplete word denoising embeddings $\mathbf{Z^*}$ are then generated by projecting $\mathbf{X}$ into the denoising matrix $\mathbf{Q_o}$ and using the non-linear activation function $\mathcal{G}$ in the following Equation~\ref{eq:transform_z}:
\begin{equation}
	\mathbf{Z^*} = \mathcal{G}(\mathbf{XQ_o})
	\label{eq:transform_z}
\end{equation}

\subsection{Loss Function}
\label{subsec:loss_func}
For each pair of term vectors $\mathbf{x_i} \in \mathbf{X}$ and $\mathbf{y_i} \in \mathbf{Y} = f(\mathbf{X,Q_c,S})$, we make use of the cosine similarity to measure the similarity between $\mathbf{x_i}$ and $\mathbf{y_i}$ as follows:
\begin{equation}
	sim(\mathbf{x_i}, \mathbf{y_i}) = \frac{\mathbf{x_i \cdot y_i}}{\mathbf{\|x_i\| \|y_i\|}}
	\label{eq:cosine-similarity}
\end{equation}
Let $\Delta$ be the difference between $sim(\mathbf{x_i}, \mathbf{x_i})$ and $sim(\mathbf{x_i}, \mathbf{y_i})$, equivalently $\Delta = 1 - sim(\mathbf{x_i}, \mathbf{y_i})$. We then optimize the objective function in Equation~\ref{eq:ObjComplete} by minimizing $\Delta$; and the same loss function is also applied to optimize the objective function in Equation~\ref{eq:ObjOvercomplete}. Training is done through Stochastic Gradient Descent with the Adadelta update rule~\cite{Zeiler:12}.

\subsection{Sparse Coding}
\label{subsec:sparse-coding}
Sparse coding is a method to represent vector representations as a sparse linear combination of elementary atoms of a given dictionary. The underlying assumption of sparse coding is that the input vectors can be reconstructed accurately as a linear combination of some basis vectors and a few number of non-zero coefficients~\cite{Olshausen/Field:96}. 

The goal is to approximate a dense vector in $\mathbb{R}^L$ by a sparse linear combination of a few columns of a matrix $\mathbf{D} \in \mathbb{R}^{L \times K}$ in which $K$ is a new vector length and the matrix $\mathbf{D}$ be called a \textit{dictionary}. Concretely, given $V$ input vectors of $L$ dimensions $\mathbf{X}=[x_1,x_2,...,x_V]$, the dictionary and sparse vectors can be formulated as the following minimization problem:
\begin{equation}
\mathop {\min }\limits_{\mathbf{D} \in \mathcal{C},\mathbf{Z} \in \mathbb{R}^{K\times{V}} }\sum\limits_{i = 1}^V \left\| {\mathbf{x_i} - \mathbf{D}{\mathbf{z}_i}} \right\|_2^2 + \lambda {\left\| {{\mathbf{z}_i}} \right\|_1}
\label{eq:sparsecode}
\end{equation}
$\mathbf{Z} = [z_1,...,z_V]$ carries the decomposition coefficients of $\mathbf{X}=[x_1,x_2,...,x_V]$; and $\lambda$ represents a scalar to control the sparsity level of $\mathbf{Z}$.
The dictionary $\mathbf{D}$ is typically learned by minimizing Equation~\ref{eq:sparsecode} over input vectors $\mathbf{X}$. In the case of overcomplete representations $\mathbf{Z}$, the vector length $K$ is typically implied as $K=\gamma L$ $(\gamma > 0)$.

In the method of overcomplete word denoising embeddings (Section~\ref{subsec:overcomplete-representation}), our approach makes use of overcomplete word embeddings $\mathbf{Z}$ as the intermediate word embeddings reconstructed by applying a sparse coding method to word embeddings $\mathbf{X}$. The overcomplete word embeddings $\mathbf{Z}$ are then utilized to optimize Equation~\ref{eq:ObjOvercomplete}. To obtain overcomplete word embeddings $\mathbf{Z}$ and dictionaries, we use the \texttt{SPAMS} package\footnote{http://spams-devel.gforge.inria.fr} to implement sparse coding for word embeddings $\mathbf{X}$ and to train the dictionaries $\mathbf{D}$.

\section{Experiments}
\label{sec:experiments}

\subsection{Experimental Settings}
\label{subsec:exper-setup}
As input word embeddings, we rely on two state-of-the-art word embeddings methods: word2vec~\cite{Mikolov:13} and GloVe~\cite{Pennington:14}. We use the \texttt{word2vec} tool\footnote{https://code.google.com/p/word2vec/} and the web corpus \textit{ENCOW14A}~\cite{Schaefer2012,Schaefer2015} which contains approximately 14.5 billion tokens, in order to train Skip-gram models with 100 and 300 dimensions. For the GloVe method, we use pre-trained vectors of 100 and 300 dimensions\footnote{http://www-nlp.stanford.edu/projects/glove/} that were trained on 6 billion words from Wikipedia and English Gigaword. The $\tanh$ function is used as the non-linear activation function in both approaches. The fixed depth of filter $T$ is set to 3; further hyperparameters are chosen as discussed in Section~\ref{subsec:hyperparameter}. To train the networks, we use the \texttt{Theano} framework~\cite{Theano} to implement our models with a mini-batch size of 100. Regularization is applied by dropouts of 0.5 and 0.2 for input and output layers (without tuning), respectively. 

\subsection{Hyperparameter Tuning}
\label{subsec:hyperparameter} 
\begin{figure*}[t]
	\centering
	\begin{subfigure}{.49\textwidth}
		\centering
		\includegraphics[width=.95\textwidth]{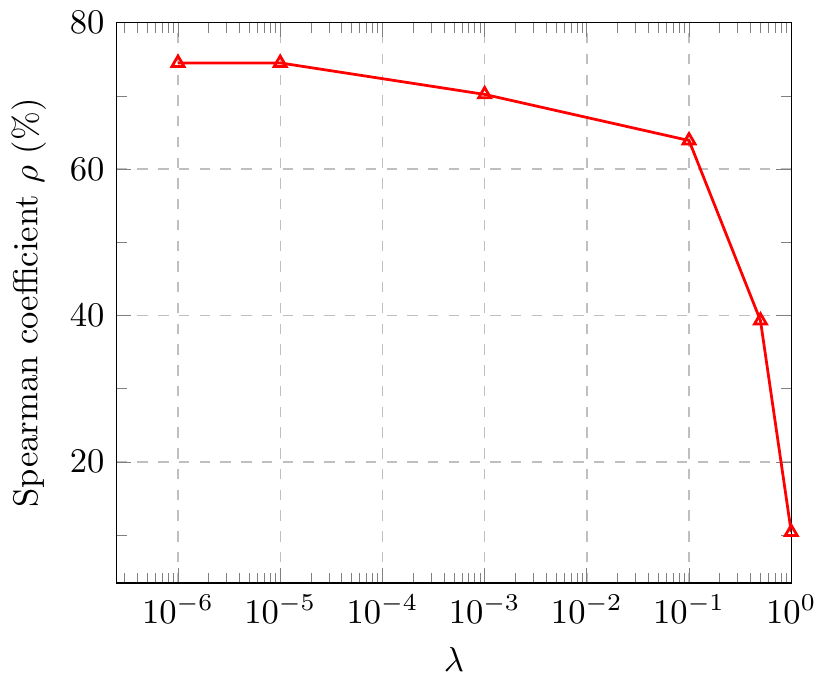} %
		\caption{Sparsity of sparse coding.}
		\label{subfig:sparsity}
	\end{subfigure}
	\begin{subfigure}{.49\textwidth}
		\centering
		\includegraphics[width=.95\textwidth]{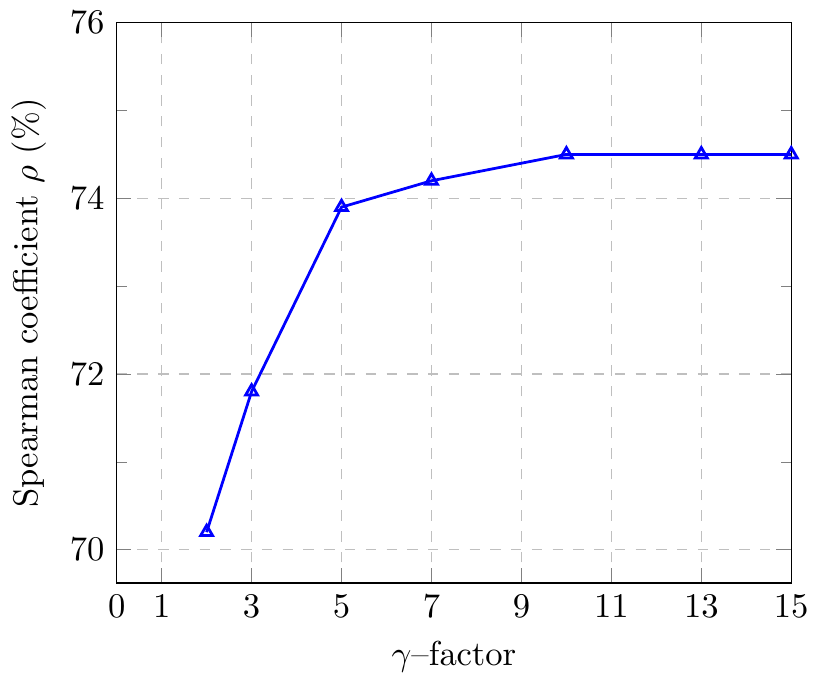} %
		\caption{Length of overcomplete vectors.}
		\label{subfig:vector-length}
	\end{subfigure}
	\caption{Illustration of hyperparameter tuning.}
	\label{fig:hyperparameter}
\end{figure*}

In both methods of denoising word embeddings, the $\ell_1$ regularization penalty $\alpha$ is set to 0.5 without tuning in Equation~\ref{eq:ObjComplete} and~\ref{eq:ObjOvercomplete}. The method of learning overcomplete word denoising embeddings relies on the mediate word embeddings $\mathbf{Z}$ to minimize the objective function in Equation~\ref{eq:ObjOvercomplete}. The sparsity of $\mathbf{Z}$ depends on the $\ell_1$ regularization $\lambda$ in Equation~\ref{eq:sparsecode}; and the length vector $K$ of $\mathbf{Z}$ is implied as $K = \gamma L$. Therefore, we aim to tune $\lambda$ and $\gamma$ such that $\mathbf{Z}$ represents the nearest approximation of the original vector representation $\mathbf{X}$. We perform a grid search on $\lambda \in \{1.0,0.5,0.1,10^{-3},10^{-6}\}$ and $\gamma \in \{2,3,5,7,10,13,15\}$, developing on the word similarity task WordSim353 (to be discussed on Section~\ref{subsec:denoising-effects}). The hyperparameter tunings are illustrated in Figures~\ref{subfig:sparsity} and~\ref{subfig:vector-length} for sparsity and overcomplete vector length tuning, respectively. In both approaches, we set $\lambda$ to $10^{-6}$ and $\gamma$ to 10 for the sparsity and length of overcomplete word embeddings.

\subsection{Effects of Word Denoising Embeddings}
\label{subsec:denoising-effects}
In this section, we quantify the effects of word denoising embeddings on three kinds of tasks: similarity and relatedness tasks, detecting synonymy, and bracketed noun phrase classification task. In comparison to the performance of word denoising embeddings, we take into account state-of-the-art word embeddings (Skip-gram and  GloVe word embeddings) as baselines. Besides, we also use the public source code\footnote{https://github.com/mfaruqui/sparse-coding} to re-implement the two methods suggested by~\newcite{Faruqui:15} which are vectors $\mathbf{A}$ (sparse overcomplete vectors) and $\mathbf{B}$ (sparse binary overcomplete vectors).

The effects of the word denoising embeddings on the tasks are shown in Table~\ref{tbl:evaluate}. The results show that the vectors $\mathbf{X^*}$ and $\mathbf{Z^*}$ outperform the original vectors $\mathbf{X, A}$ and $\mathbf{B}$, except for the NP task, in which the vectors $\mathbf{B}$ based on the 300-dimensional GloVe vectors are best. The effect of the vectors $\mathbf{Z^*}$ is slightly less impressive, when compared to the overcomplete vectors $\mathbf{X^*}$. The overcomplete word embeddings $\mathbf{Z}$ strongly differ from the word embeddings $\mathbf{X}$; hence, the denoising is affected. However, the performance of the vectors $\mathbf{Z^*}$ still outperforms the original vectors $\mathbf{X, A}$ and $\mathbf{B}$ after the denoising process.
\begin{table}[]
\centering
\resizebox{\textwidth}{!}{
\begin{tabular}{|c|l|ccccc|cc|c|}
\hline
\multicolumn{2}{|c|}{$\textbf{Vectors}$}         & \begin{tabular}[c]{@{}c@{}}$\textbf{Simlex-999}$\\ Corr.\end{tabular} & \begin{tabular}[c]{@{}c@{}}$\textbf{MEN}$\\ Corr.\end{tabular} & \begin{tabular}[c]{@{}c@{}}$\textbf{WS353}$\\ Corr.\end{tabular} & \begin{tabular}[c]{@{}c@{}}$\textbf{WS353-SIM}$\\ Corr.\end{tabular} & \begin{tabular}[c]{@{}c@{}}$\textbf{WS353-REL}$\\ Corr.\end{tabular} & \begin{tabular}[c]{@{}c@{}}$\textbf{ESL}$\\ Acc.\end{tabular} & \begin{tabular}[c]{@{}c@{}}$\textbf{TOEFL}$\\ Acc.\end{tabular} & \begin{tabular}[c]{@{}c@{}}$\textbf{NP}$\\ Acc.\end{tabular} \\ \hline
                                & $\mathbf{X}$   & 33.7                                                                  & 72.9                                                           & 69.7                                                             & 74.5                                                                 & 65.5                                                                 & 48.9                                                          & 62.0                                                            & 72.8                                                         \\
                                & $\mathbf{X^*}$ & 33.2                                                                  & 72.8                                                           & 70.6                                                             & 74.8                                                                 & 66.0                                                                 & 53.0                                                          & \textbf{64.5}                                                   & 78.5                                                         \\
SG-100                          & $\mathbf{Z^*}$ & \textbf{35.9}                                                         & \textbf{74.4}                                                  & \textbf{71.2}                                                    & \textbf{75.2}                                                        & \textbf{68.1}                                                        & 53.0                                                          & 62.0                                                            & \textbf{79.1}                                                \\
\multicolumn{1}{|l|}{}          & $\mathbf{A}$   & 32.5                                                                  & 69.8                                                           & 65.5                                                             & 69.5                                                                 & 60.2                                                                 & \textbf{55.1}                                                 & 51.8                                                            & 78.8                                                         \\
\multicolumn{1}{|l|}{}          & $\mathbf{B}$   & 31.9                                                                  & 70.4                                                           & 65.8                                                             & 72.6                                                                 & 62.2                                                                 & 53.0                                                          & 58.2                                                            & 74.1                                                         \\ \hline
                                & $\mathbf{X}$   & 36.1                                                                  & 74.7                                                           & 71.0                                                             & 75.9                                                                 & 66.1                                                                 & \textbf{59.1}                                                 & 72.1                                                            & 77.9                                                         \\
                                & $\mathbf{X^*}$ & \textbf{37.1}                                                         & \textbf{75.8}                                                  & \textbf{71.8}                                                    & \textbf{76.4}                                                        & \textbf{66.9}                                                        & \textbf{59.1}                                                 & 74.6                                                            & \textbf{79.3}                                                \\
SG-300                          & $\mathbf{Z^*}$ & 36.5                                                                  & 75.0                                                           & 70.6                                                             & \textbf{76.4}                                                        & 64.4                                                                 & 57.1                                                          & \textbf{77.2}                                                   & 78.6                                                         \\
\multicolumn{1}{|l|}{}          & $\mathbf{A}$   & 32.9                                                                  & 72.4                                                           & 67.5                                                             & 71.9                                                                 & 63.4                                                                 & 53.0                                                          & 65.8                                                            & 78.3                                                         \\
\multicolumn{1}{|l|}{}          & $\mathbf{B}$   & 32.7                                                                  & 71.2                                                           & 63.3                                                             & 68.7                                                                 & 56.2                                                                 & 51.0                                                          & 70.8                                                            & 78.6                                                         \\ \hline
                                & $\mathbf{X}$   & 29.7                                                                  & 69.3                                                           & 52.9                                                             & 60.3                                                                 & 49.5                                                                 & 46.9                                                          & 82.2                                                            & 76.4                                                         \\
                                & $\mathbf{X^*}$ & \textbf{31.7}                                                         & \textbf{70.9}                                                  & \textbf{58.0}                                                    & \textbf{63.8}                                                        & \textbf{57.3}                                                        & 53.0                                                          & \textbf{88.6}                                                   & \textbf{77.4}                                                \\
GloVe-100                       & $\mathbf{Z^*}$ & 30.0                                                                  & \textbf{70.9}                                                  & 56.0                                                             & 62.8                                                                 & 53.8                                                                 & \textbf{57.0}                                                 & 81.0                                                            & 77.3                                                         \\
\multicolumn{1}{|l|}{}          & $\mathbf{A}$   & 30.7                                                                  & 70.7                                                           & 54.9                                                             & 62.2                                                                 & 51.2                                                                 & 55.1                                                          & 78.4                                                            & 77.1                                                         \\
\multicolumn{1}{|l|}{}          & $\mathbf{B}$   & 31.0                                                                  & 69.2                                                           & 57.3                                                             & 62.3                                                                 & 53.7                                                                 & 46.9                                                          & 73.4                                                            & 76.4                                                         \\ \hline
                                & $\mathbf{X}$   & 37.0                                                                  & 74.8                                                           & 60.5                                                             & 66.3                                                                 & 57.2                                                                 & \textbf{61.2}                                                 & 89.8                                                            & 74.3                                                         \\
\multicolumn{1}{|l|}{}          & $\mathbf{X^*}$ & \textbf{40.2}                                                         & \textbf{76.8}                                                  & \textbf{64.9}                                                    & \textbf{69.8}                                                        & \textbf{62.0}                                                        & \textbf{61.2}                                                 & \textbf{92.4}                                                   & 76.3                                                         \\
\multicolumn{1}{|l|}{GloVe-300} & $\mathbf{Z^*}$ & 39.0                                                                  & 75.2                                                           & 63.0                                                             & 67.9                                                                 & 59.7                                                                 & 57.1                                                          & 86.0                                                            & 75.7                                                         \\
\multicolumn{1}{|l|}{}          & $\mathbf{A}$   & 36.7                                                                  & 74.1                                                           & 61.5                                                             & 67.7                                                                 & 57.8                                                                 & 55.1                                                          & 87.3                                                            & 79.9                                                         \\
\multicolumn{1}{|l|}{}          & $\mathbf{B}$   & 33.1                                                                  & 70.2                                                           & 57.0                                                             & 62.2                                                                 & 53.0                                                                 & 51.0                                                          & 91.4                                                            & \textbf{80.0}                                                \\ \hline
\end{tabular}
}
\caption{Effects of word denoising embeddings. Vectors $\mathbf{X}$ represent the baselines; vectors $\mathbf{A}$ and $\mathbf{B}$ were suggested by~\newcite{Faruqui:15}; the vector length $\mathbf{Z^*}$ is equal to 10 times of vector length $\mathbf{X}$.}
\label{tbl:evaluate}
\vspace{+5mm}
\end{table}

\subsubsection{Relatedness and Similarity Tasks} For the relatedness task, we use two kinds of datasets: MEN~\cite{Bruni:14} consists of 3000 word pairs comprising 656 nouns, 57 adjectives and 38 verbs. The WordSim-353 relatedness dataset~\cite{Finkelstein:01} contains 252 word pairs. Concerning the similarity tasks, we evaluate the denoising vectors again on two kinds of datasets: \textit{SimLex-999}~\cite{Hill2015} contains 999 word pairs including 666 noun, 222 verb and 111 adjective pairs. The WordSim-353 similarity dataset consists of 203 word pairs. In addition, we evaluate our denoising vectors on the WordSim-353 dataset which contains 353 pairs for both similarity and relatedness relations. We calculate cosine similarity between the vectors of two words forming a test pair, and report the Spearman rank-order correlation coefficient $\rho$~\cite{Siegel/Castellan:88} against the respective gold standards of human ratings.


\subsubsection{Synonymy}
We evaluate on 80 TOEFL (Test of English as a Foreign Language) synonym questions~\cite{Landauer/Dutnais:97} and 50 ESL (English as a Second Language) questions~\cite{Turney:01}. The first dataset represents a subset of 80 multiple-choice synonym questions from the TOEFL test: a word is paired with four options, one of which is a valid synonym. The second dataset contains 50 multiple-choice synonym questions, and the goal is to choose a valid synonym from four options. For each question, we compute the cosine similarity between the target word and the four candidates. The suggested answer is the candidate with the highest cosine score. We use accuracy to evaluate the performance.

\subsubsection{Phrase parsing as Classification}
 \newcite{Lazaridou:13} introduced a dataset of noun phrases (NP) in which each NP consists of three elements: the first element is either an adjective or a noun, and the other elements are all nouns. For a given NP (such as \textit{blood pressure medicine}), the task is to predict whether it is a left-bracketed NP, e.g., \textit{(blood pressure) medicine}, or a right-bracketed NP, e.g., \textit{blood (pressure medicine)}.

The dataset contains 2227 noun phrases split into 10 folds. For each NP, we use the average of word vectors as features to feed into the classifier by tuning the hyperparameters ($w_1$, $w_2$ and $w_3$) for each element ($e_1$, $e_2$ and $e_3$) within the NP: $\vec{e}_{NP} = \frac{1}{3}(w_1\vec{e}_1 + w_2\vec{e}_2 + w_3\vec{e}_3)$. We then employ the classification of the NPs by using a Support Vector Machine (SVM) with Radial Basis Function kernel. The classifier is tuned on the first fold, and cross-validation accuracy is reported on the nine remaining folds.

\subsection{Effects of Filter Depth}
\label{subsec:depth-effects}
As mentioned above, the architecture of the filter $f$ is a feed-forward network with a fixed depth $T$. For each stage $T$, the filter $f$ attempts to reduce the noise within input vectors by approximating these vectors based on vectors of a previous stage $T-1$. In order to investigate the effects of each stage $T$, we use pre-trained GloVe vectors with 100 dimensions to evaluate the denoising performance of the vectors on detecting synonymy in the TOEFL dataset across several stages of $T$.
\begin{figure}[h]
	\centering
	\includegraphics[width=.4\textwidth]{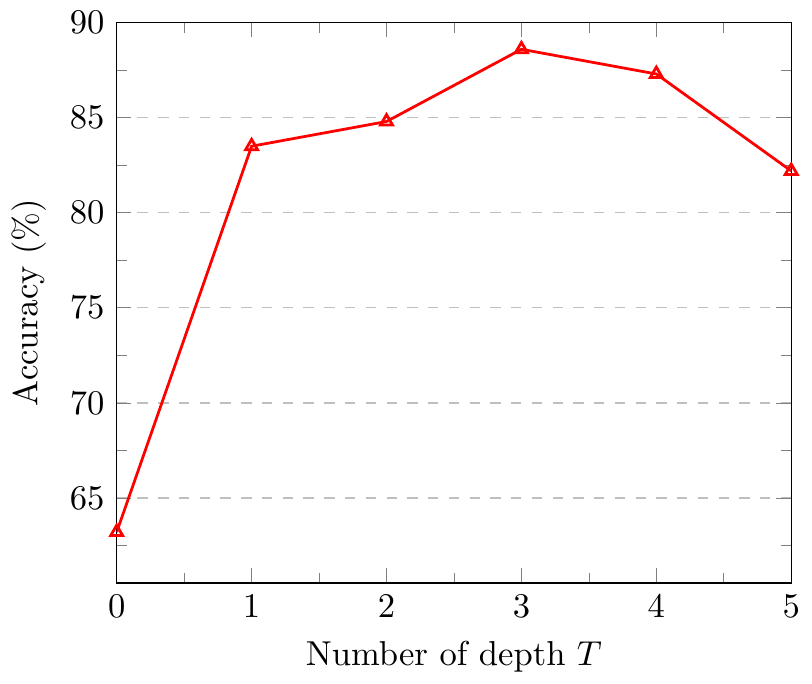} %
	\caption{Effects of the filter with depth $T$ on filtering noise.}
	\label{fig:depth}
        \vspace{+5mm}
\end{figure}

The results are presented in Figure~\ref{fig:depth}. The accuracy of synonymy detection increases sharply from 63.2\% to 88.6\% according to the number of stages $T$ from 0 to 3. However, the denoising performance of vectors falls with the number of stages $T>3$. This evaluation shows that the filter $f$ with a consistently fixed depth $T$ can be trained to efficiently filter noise for word embeddings. In other words, the number of stages $T$ exceeds a consistent number $T$ (with $T > 3$ in our case), leading to the loss of salient information in the vectors.


\section{Conclusion}
\label{sec:conclusion}
To the best of our knowledge, we are the first to work on filtering noise in word embeddings. In this paper, we have presented two novel models to improve word embeddings by reducing noise in state-of-the-art word embeddings models. The underlying idea in our models was to make use of a deep feed-forward neural network filter to reduce noise. The first model generated complete word denoising embeddings; the second model yielded overcomplete word denoising embeddings. We demonstrated that the word denoising embeddings outperform the originally state-of-the-art word embeddings on several benchmark tasks.
\section*{Acknowledgements}
The research was supported by the Ministry of Education and Training of the Socialist Republic of Vietnam (Scholarship 977/QD-BGDDT; Nguyen Kim Anh) and the DFG Heisenberg Fellowship SCHU-2580/1 (Sabine Schulte im Walde). It is also a collaboration between project D12 and project A8 in the DFG Collaborative Research Centre SFB 732.
\bibliographystyle{acl}
\bibliography{denoising-embeddings}

\begin{thebibliography}{}

\bibitem[\protect\citename{Adel and Sch{\"{u}}tze}2014]{Adel/Schuetze:14}
Heike Adel and Hinrich Sch{\"{u}}tze.
\newblock 2014.
\newblock Using mined coreference chains as a resource for a semantic task.
\newblock In {\em Proceedings of the 2014 Conference on Empirical Methods in
  Natural Language Processing (EMNLP)}, pages 1447--1452, Doha, Qatar.

\bibitem[\protect\citename{Bruni \bgroup et al.\egroup }2014]{Bruni:14}
Elia Bruni, Nam{-}Khanh Tran, and Marco Baroni.
\newblock 2014.
\newblock Multimodal distributional semantics.
\newblock {\em Journal of Artifical Intelligence Research {(JAIR)}}, 49:1--47.

\bibitem[\protect\citename{Faruqui \bgroup et al.\egroup }2015]{Faruqui:15}
Manaal Faruqui, Yulia Tsvetkov, Dani Yogatama, Chris Dyer, and Noah~A. Smith.
\newblock 2015.
\newblock Sparse overcomplete word vector representations.
\newblock In {\em Proceedings of the 53rd Annual Meeting of the Association for
  Computational Linguistics (ACL)}, pages 1491–--1500, Beijing, China.

\bibitem[\protect\citename{Finkelstein \bgroup et al.\egroup
  }2001]{Finkelstein:01}
Lev Finkelstein, Evgeniy Gabrilovich, Yossi Matias, Ehud Rivlin, Zach Solan,
  Gadi Wolfman, and Eytan Ruppin.
\newblock 2001.
\newblock Placing search in context: The concept revisited.
\newblock In {\em Proceedings of the 10th International Conference on the World
  Wide Web}, pages 406--414.

\bibitem[\protect\citename{Gregor and LeCun}2010]{Gregor/LeCun:10}
Karol Gregor and Yann LeCun.
\newblock 2010.
\newblock Learning fast approximations of sparse coding.
\newblock In {\em Proceedings of the 27th International Conference on Machine
  Learning (ICML), Haifa, Israel}, pages 399--406.

\bibitem[\protect\citename{Harris}1954]{Harris1954}
Zellig~S. Harris.
\newblock 1954.
\newblock Distributional structure.
\newblock {\em Word}, 10(23):146--162.

\bibitem[\protect\citename{Hill \bgroup et al.\egroup }2015]{Hill2015}
Felix Hill, Roi Reichart, and Anna Korhonen.
\newblock 2015.
\newblock Simlex-999: Evaluating semantic models with (genuine) similarity
  estimation.
\newblock {\em Computational Linguistics}, 41(4):665--695.

\bibitem[\protect\citename{Kim}2014]{Kim:14}
Yoon Kim.
\newblock 2014.
\newblock Convolutional neural networks for sentence classification.
\newblock In {\em Proceedings of the 2014 Conference on Empirical Methods in
  Natural Language Processing ({EMNLP})}, pages 1746--1751, Doha, Qatar.

\bibitem[\protect\citename{Landauer and Dumais}1997]{Landauer/Dutnais:97}
Thomas~K. Landauer and Susan~T. Dumais.
\newblock 1997.
\newblock A solution to {P}lato’s problem: The latent semantic analysis
  theory of acquisition, induction, and representation of knowledge.
\newblock {\em Psychological Review}, 104(2):211--240.

\bibitem[\protect\citename{Lazaridou \bgroup et al.\egroup }2013]{Lazaridou:13}
Angeliki Lazaridou, Eva~Maria Vecchi, and Marco Baroni.
\newblock 2013.
\newblock Fish transporters and miracle homes: How compositional distributional
  semantics can help {NP} parsing.
\newblock In {\em Proceedings of the 2014 Conference on Empirical Methods in
  Natural Language Processing (EMNLP)}, pages 1908–--1913, Doha, Qatar.

\bibitem[\protect\citename{Mikolov \bgroup et al.\egroup }2013]{Mikolov:13}
Tomas Mikolov, Wen-tau Yih, and Geoffrey Zweig.
\newblock 2013.
\newblock Linguistic regularities in continuous space word representations.
\newblock In {\em Proceedings of the 2013 Conference of the North American
  Chapter of the Association for Computational Linguistics: Human Language
  Technologies (NAACL)}, pages 746--751, Atlanta, Georgia.

\bibitem[\protect\citename{Nguyen \bgroup et al.\egroup }2016]{Nguyen:16}
Kim~Anh Nguyen, Sabine Schulte~im Walde, and Ngoc~Thang Vu.
\newblock 2016.
\newblock Integrating distributional lexical contrast into word embeddings for
  antonym-synonym distinction.
\newblock In {\em Proceedings of the 54th Annual Meeting of the Association for
  Computational Linguistics (ACL)}, pages 454–--459, Berlin, Germany.

\bibitem[\protect\citename{Olshausen and Field}1996]{Olshausen/Field:96}
Bruno~A. Olshausen and David~J. Field.
\newblock 1996.
\newblock Emergence of simple-cell receptive field properties by learning a
  sparse code for natural images.
\newblock {\em Nature}, 381(6583):607--609.

\bibitem[\protect\citename{Pennington \bgroup et al.\egroup
  }2014]{Pennington:14}
Jeffrey Pennington, Richard Socher, and Christopher~D. Manning.
\newblock 2014.
\newblock Glove: Global vectors for word representation.
\newblock In {\em Proceedings of the 2014 Conference on Empirical Methods in
  Natural Language Processing ({EMNLP})}, pages 1532--1543, Doha, Qatar.

\bibitem[\protect\citename{Pham \bgroup et al.\egroup }2015]{Pham:15}
Nghia~The Pham, Angeliki Lazaridou, and Marco Baroni.
\newblock 2015.
\newblock A multitask objective to inject lexical contrast into distributional
  semantics.
\newblock In {\em Proceedings of the 53rd Annual Meeting of the Association for
  Computational Linguistics (ACL)}, pages 21--26, Beijing, China.

\bibitem[\protect\citename{Sch\"afer and Bildhauer}2012]{Schaefer2012}
Roland Sch\"afer and Felix Bildhauer.
\newblock 2012.
\newblock Building large corpora from the web using a new efficient tool chain.
\newblock In {\em Proceedings of the 8th International Conference on Language
  Resources and Evaluation}, pages 486--493, Istanbul, Turkey.

\bibitem[\protect\citename{Sch\"afer}2015]{Schaefer2015}
Roland Sch\"afer.
\newblock 2015.
\newblock Processing and querying large web corpora with the {COW14}
  architecture.
\newblock In {\em Proceedings of the 3rd Workshop on Challenges in the
  Management of Large Corpora}, pages 28--34, Mannheim, Germany.

\bibitem[\protect\citename{Siegel and Castellan}1988]{Siegel/Castellan:88}
Sidney Siegel and N.~John Castellan.
\newblock 1988.
\newblock {\em Nonparametric Statistics for the Behavioral Sciences}.
\newblock McGraw-Hill, Boston, MA.

\bibitem[\protect\citename{Szlam \bgroup et al.\egroup }2011]{Szlam:11}
Arthur~D. Szlam, Karol Gregor, and Yann~L. Cun.
\newblock 2011.
\newblock Structured sparse coding via lateral inhibition.
\newblock {\em Advances in Neural Information Processing Systems (NIPS)},
  24:1116--1124.

\bibitem[\protect\citename{{Theano Development Team}}2016]{Theano}
{Theano Development Team}.
\newblock 2016.
\newblock {Theano: A {Python} framework for fast computation of mathematical
  expressions}.
\newblock {\em arXiv e-prints}, abs/1605.02688.

\bibitem[\protect\citename{Turney}2001]{Turney:01}
Peter~D. Turney.
\newblock 2001.
\newblock Mining the web for synonyms: {PMI-IR} versus {LSA} on {TOEFL}.
\newblock In {\em Proceedings of the 12th European Conference on Machine
  Learning (ECML)}, pages 491--502.

\bibitem[\protect\citename{Zeiler}2012]{Zeiler:12}
Matthew~D. Zeiler.
\newblock 2012.
\newblock {ADADELTA:} an adaptive learning rate method.
\newblock {\em CoRR}, abs/1212.5701.

\end{thebibliography}
\end{document}